# MSDC-Net: Multi-Scale Dense and Contextual Networks for Automated Disparity Map for Stereo Matching

Zhibo Rao[a], Mingyi He*[a], Yuchao Dai[a], Zhidong Zhu[a], Bo Li[a], Renjie He[a,b]

[a] Northwestern Polytechnical University, Xi'an 710129, China

[b] Nanyang Technological University, 639798, Singapore

**ABSTRACT**

Disparity prediction from stereo images is essential to computer vision applications including autonomous driving, 3D model reconstruction, and object detection. To predict accurate disparity map, we propose a novel deep learning architecture for detecting the disparity map from a rectified pair of stereo images, called MSDC-Net. Our MSDC-Net contains two modules: multi-scale fusion 2D convolution and multi-scale residual 3D convolution modules. The multi-scale fusion 2D convolution module exploits the potential multi-scale features, which extracts and fuses the different scale features by Dense-Net. The multi-scale residual 3D convolution module learns the different scale geometry context from the cost volume which aggregated by the multi-scale fusion 2D convolution module. Experimental results on Scene Flow and KITTI datasets demonstrate that our MSDC-Net significantly outperforms other approaches in the non-occluded region.

## 1. INTRODUCTION

Stereo disparity estimation aims at predicting the disparity $d$ from a pair of stereo images, which is an essential intermediate component toward 3D scene reconstruction and understanding. For example, the disparity map information can benefit tasks such as autonomous driving for vehicles [1, 2], object detection and recognition [3, 4], and 3D model reconstruction [5-7].

In general, traditional stereo matching methods care more about how to accurately compute the matching cost and how to apply local or global information to refine the disparity map [8-10]. C. Rhemann et al. replaced the cost volume by considering cost aggregation methods as joint filtering. Their method proved that simple linear image filters such as a box or gaussian filter could even be used for cost aggregation [11]. K. Zhang et al. took advantage of cross-scale cost aggregation to optimize the cost volume. It showed cross-scale framework is useful and leads to significant improvements [12]. K. Zhang et al. employed an area-based local stereo matching algorithm for all image regions to evaluate disparity map, efficient approach that finding the matching points of given points within a predefined support window [13]. H. Simon et al. proposed a semi-global matching approach based on the coarse-to-fine (CTF) strategy to accelerate convergence and to avoid unexpected local minima [14]. The traditional methods tend to be highly explanatory and adaptable.

Recently, deep learning has made considerable achievements in understanding semantics from the raw data in matching corresponding points. Compared with the conventional methods, deep learning based methods are capable of making significant improvements in both precision and speed [15-18]. GC-Net employed the hierarchical 3D convolutions architecture to learn context from the cost volume which concatenated by each unary feature [17]. SsSM-Net proposed a novel training loss to exploit the loop constraint in image warping and to handle the texture-less areas, leaving it can self-improve by adapting itself to new imageries [19]. SGM-Net utilized the penalties estimation method to control the smoothness and discontinuity of the disparity map [20]. PSM-Net [18] exploited global context information by spatial pyramid pooling (SPP) [21, 22] and dilated convolution architectures [23]. Mayer et al. introduced two end-to-end networks for disparity estimation (Disp-Net) and optical flow (Flow-Net). They also created a large synthetic dataset called scene flow to improve the state-of-the-art [24]. S. Zagoruyko et al. trained a non-learned cost aggregation and regularization combined deep network to match $5 \times 5$ image patches. It shows that multiple neural network architectures specifically adapt to the stereo matching task [25]. The main idea of these methods is to extract the features and learn the context information from stereo image pairs, thus improves the accuracy of disparity estimation.

In this work, we propose a novel multi-scale dense and contextual stereo matching network (MSDC-Net) to exploit global context information in stereo matching effectively. We design a multi-scale fusion 2D convolution module to improve the global context understanding ability by extracting the cross-scale feature. Moreover, we redesign the 3D convolution module from the GC-Net and introduce the multi-scale residual 3D convolution, which improves the utilization of global context information. The experimental results prove that the proposed architecture outperforms GC-Net in learning global context.

Our main contributions can be summarized as:

(1) We propose an end-to-end learning framework for stereo matching without any post-processing.

(2) We design a multi-scale fusion 2D convolution module for incorporating global context information from image pairs.

(3) We redesign a multi-scale residual 3D convolution to learn the regional support of context information in cross-scale cost volume.

## 2. OUR METHODS

In this section, we present multi-scale dense and contextual network (MSDC-Net) in detail. The network architecture is illustrated in Figure 1. Our model consists of four steps: multi-scale features extraction and fusion, cost volume construction, feature matching, and disparity map regression. First, the multi-scale fusion 2D convolution module is applied to extract and fuse the multi-scale features. Then, feature pairs are alternated to aggregate cost volume as shown in section 2.2. After that, the matching features are learnt and the size of features volume is recovered by the multi-scale residual 3D convolution module. Finally, the disparity map is obtained by features volume regression. The implementation detail is described in the following subsections, respectively.

* Mingyi He, corresponding author, Email: myhe@nwpu.edu.cn

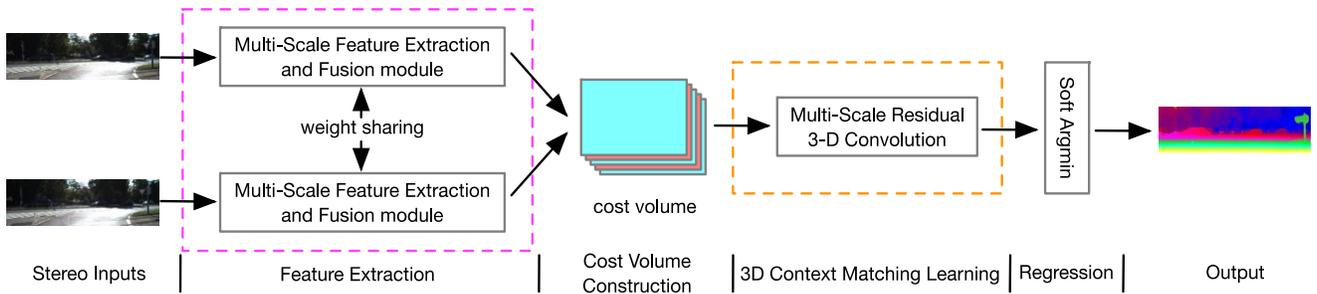

Figure 1: Our end-to-end deep stereo regression architecture: MSDC-Net (Multi-Scale Dense and Contextual architecture).

## 2.1 Feature Extraction and Fusion by Multi-Scale Fusion 2D Convolution Module

To get a robust descriptor which could represent the ambiguities in the photometric region and can incorporate local context, it is common to use a feature representation to capture local context [26]. We could use the deep feature representation to incorporate hierarchical context information by the Dense-Net [26].

In our model, we design a multi-scale fusion 2D convolution module through a series of 2D convolutional operations as shown in Fig. 2. The basic feature number is 32, and each convolutional layer is followed by a BN layer and a ReLU layer. The multi-scale fusion 2D convolution module contains two parts: different scale feature extraction (DSFE) and multi-scale features fusion (MSFF). DSFE is applied to extract the features with different size from image pairs. This part owns the 51 convolution layers with the different convolution filters. To reduce the calculation complexity, we initially adopt the $5 \times 5$ convolutional filter with the stride of two to subsample the stereo image pairs. Following this layer, we apply the dense block consisting of 16 convolutional layers with $3 \times 3$ convolu-tional filters and direct connections between four convolutional layers. Thus, we could obtain the feature size about $1/2H \times 1/2W$. In the same way, we could get the feature size about $1/4H \times 1/4W$ and $1/8H \times 1/8W$, and padding the feature size to $1/2H \times 1/2W$. Then we concatenate the different size features to obtain the aggregating features volume.

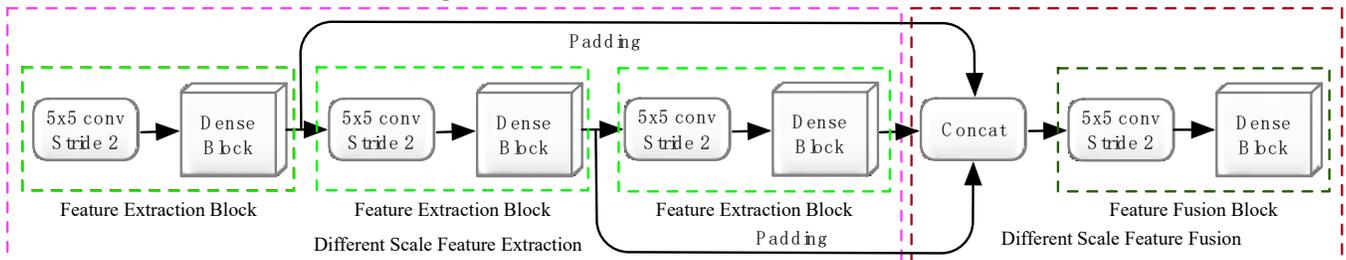

Figure 2: The multi-scale fusion 2D convolution module. The module contains two parts: different scale feature extraction and multi-scale features fusion.

Multi-scale feature fusion part fuses the aggregated features volume to form a cost volume. It has 18 convolution layers with the different filters. To avoid losing the critical information and fusing the aggregating features volume, we adopt 128 convolutional filters with the size $5 \times 5$ and the stride of two to subsample the aggregating features volume. Then the dense block fuses these features. We obtain the unary features by passing stereo images with the same parameter.

## 2.2 Cost Volume Construction by Alternate Feature Pairs

Similar to the conventional stereo matching algorithms, we construct a four-dimensional cost volume ($height \times width \times disparity \times feature$) by concatenating the fusion features at each disparity level as shown in Figure 3. Specifically, the stereo matching cost is computed using the deep unary features of stereo image pairs to preserve prior knowledge of stereo vision.

## 2.3 Feature Matching by Multi-Scale Residual 3D Convolution Module

Given the fusion feature assembled cost volume, we would like to learn the matching cost at each candidate disparity from the different size unary feature and the regularization from the local context. The encoder-decoder networks often cost a vast amount of calculation and very difficult to train. Thus, we re-design the 3D convolution module which could better learn the context of tiny objects and improve the efficiency of the learning process.

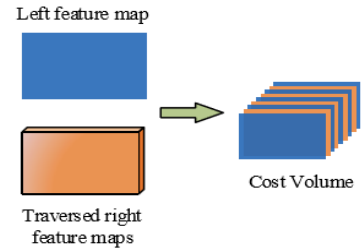

Figure 3: Cost Volume Construction. The blue rectangle represents a fusion feature map of the left image. The orange cube represents the set of traversed fusion features of the right image from 0 to the disparity $1/4D$.

To exploit the relationship between disparity, height, and width of image pairs, we present the multi-scale residual 3D convolution module, which contains two parts: multi-scale residual feature matching and scale recovery as shown in Figure 4. In the module, the basic feature number is 32.

The multi-scale feature matching part is applied to match the geometry features from cost volume. This part has four levels, and there are residual 3D convolution layers between each subsampling. When subsampling, the features number will become double.

When up-sampling, the features number will decrease one time. On this basis, we pass the features information between the same level 3D convolution layer to avoid losing the critical information.

After this part, we obtain the matching feature but in a low resolution $1/4H \times 1/4W \times 1/4D$. Therefore, we must recover scale to get the final feature volume.

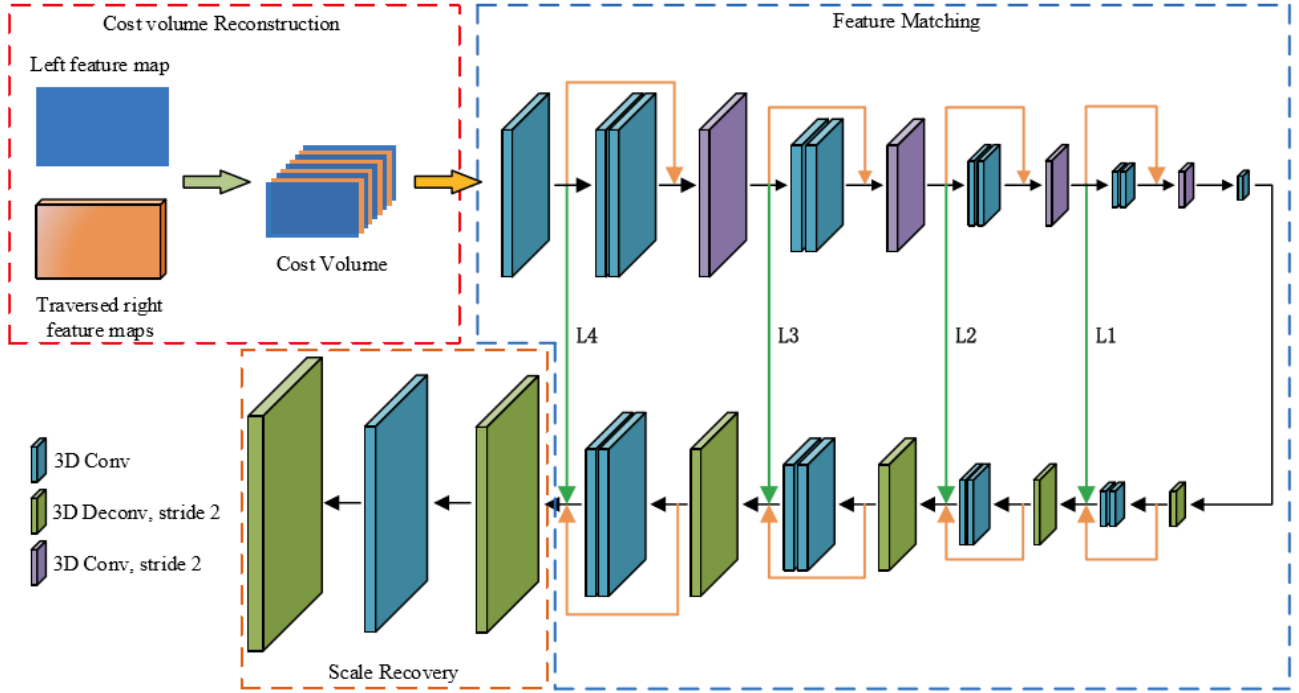

**Figure 4: The multi-scale residual 3D convolution module.** The module consists of two main parts: feature matching and scale recovery.

The scale recovery part is applied to recover the size of the input image. In this part, we customize a simple strategy to recovery scale. The output of the scale recovery part is a final feature volume with size $H \times W \times D$.

## 2.4 Disparity Map Regression by Soft argMin

Compared with the classification-based matching method, the disparity map obtained by regression is more robust and effective. Thus, we predict the disparity map by passing an argmin operation. First, we get the probability of each disparity value $d$ which can be calculated from a final feature volume $c$ via the softmax operation $\sigma(\cdot)$. Second, we obtain the predicted disparity $\hat{d}$ which can be calculated as the sum of each disparity $d$ weighted, as

$$\hat{d} = \sum_{d=0}^{D_{max}} d \times \sigma(-c_d) \qquad (1)$$

where $c$ represents the final cost volume with size $height \times width \times disparity$, and $\sigma(\cdot)$ represents the softmax operation.

## 2.5 Loss Function

The overall consideration based on the stereo matching research, we think the absolute error value of predicted result and ground-truth should be used in the different $Loss$ function. Compared to $L_2$ loss function, the $L_1$ loss is widely used in object detection because of its robustness and low sensitivity to outliers. To fit in with the stereo matching task, the $L_1$ loss function is redefined as:

$$L(d, \hat{d}_i) = \frac{1}{N} \sum_{i=1}^{N} Smooth_{L_1}(d - \hat{d}_i) \qquad (2)$$

In which

$$Smooth_{L_1}(x) = \begin{cases} \frac{1}{3}x^2, if\ |x| < 3. \\ |x|, otherwise. \end{cases} \qquad (3)$$

where $N$ is the total number of labeled pixels where the value is not 0, $d$ is the ground-true disparity, and $\hat{d}_i$ is the predicted disparity. In the loss function, we set three as the critical point. Because we hope when the predicted pixels value less than three, the influence of the points more less impact on the network, and three as the critical point is the error which could be accepted.

## 3. EXPERIMENTS

In this section, the performance of the proposed method is evaluated on two widely used stereo datasets: Scene Flow [24] and KITTI [27]. We firstly show our experiment details on the training process. After that, we discuss the different parameters of our model and justify a series of our design models in the various parameters. Finally, we compare the performance of our method with the state-of-art on the KITTI stereo dataset.

### 3.1 Experimental parameters

The proposed MSDC-Net is implemented using Tensorflow. All models were end-to-end trained by Adam Optimizer with a constant learning rate of $1 \times 10^3$ and $\beta_1 = 0.9$, $\beta_2 = 0.999$. Color normalization is performed on each image to ensure the pixel intensities ranged from 0 to 1. To increase the sample and adapt the training requirements of networks, the input images are randomly cropped to size $H = 256$ and $W = 512$ from a pair of normalized stereo images. The maximum disparity is set to $D = 192$. We trained our model on four Nvidia 1080Ti GPUs with the batch size of 8. The training process took 50 epochs for Scene Flow dataset and 1000 epochs for KITTI stereo dataset.

## 3.2 Model Design Analysis

To verify the effectiveness of our design, we present an ablation study to compare a series of different model variants. We apply the Scene Flow dataset [24] for the experiments, which contains 35,454 training and 4,370 testing images with $H = 540$ and $W = 960$. As shown in [17] and [18], the large dataset used to train the model without over-fitting, it could help to evaluate the model correctly. Moreover, the Scene Flow dataset has dense ground truth and removes any discrepancies which caused by wrong labels. To evaluate different model variants, we first train each model for 50 epochs to obtain the models, then verify the models from the test images, the result as shown in Table 1 and Figure 5.

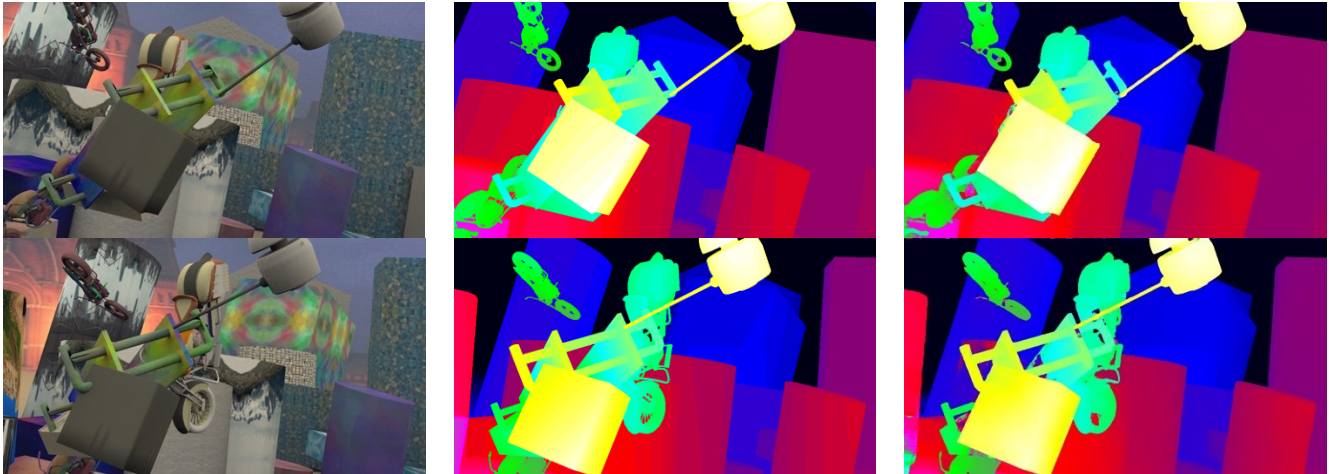

**Figure 5: Scene flow test data qualitative results.** From left: left stereo input image, ground-truth disparity map, disparity prediction.

**Table 1: Comparison on different model variants on the synthetic Scene Flow dataset** [24].

| Model Type | Error Rate (%) | | | Error | | Param. | Time (ms) |
| --- | --- | --- | --- | --- | --- | --- | --- |
| | >1px | >3px | >5px | MAE | RMS | | |
| Single scale 2D and 3D conv (replace conv layers 1-97) | 28.7 | 18.2 | 16.4 | 7.34 | 24.8 | 4.4M | 0.76 |
| Single scale 2D conv (replace 2D conv layers 1-68) w Multi-Scale Residual 3D conv | 14.9 | 9.5 | 8.1 | 3.6 | 17.9 | 4.4M | 0.62 |
| Single scale 3D conv (replace 3D conv layers 69-97) w Multi-Scale Fusion 2D conv | 15.8 | 9.2 | 7.4 | 3.8 | 16.2 | 4.6M | 0.74 |
| MSDC-Net | **11.6** | **8.7** | **6.4** | **1.6** | **11.3** | **4.6M** | 0.75 |

As shown in Table 1 and Figure 5, the multi-scale fusion 2D convolution and multi-scale residual 3D convolution architectures perform excellent performance and significantly outperform single scale 2D and 3D convolution architectures.

## 3.3 Experiment with KITTI

In order to evaluate the performance of our model with the state-of-art methods published recently, we fine-tune the model which pre-trained on Scene Flow for a further 1000 epochs on KITTI 2012 and 2015 respectively. The KITTI 2012 and 2015 are real-world datasets with challenging and varied road scene, which contain 194 training stereo image pairs in the KITTI 2012 and 200 training stereo image pairs in the KITTI 2015 with sparse ground-truth disparities. Moreover, the datasets prepare another 194 testing image pairs in the KITTI 2012 and 200 testing image pairs in the KITTI 2015 without ground-truth disparities. In order to prevent our model form over-fitting on the very small KITTI training dataset, we divided the whole training data into a training set 80% and a validation set 20%. We show the representative results of our method and comparison the state-of-art methods in Figure 6 and 7. In addition, we evaluate the performance of our model on KITTI 2012 and 2015 testing datasets respectively in Table 2 and 3.

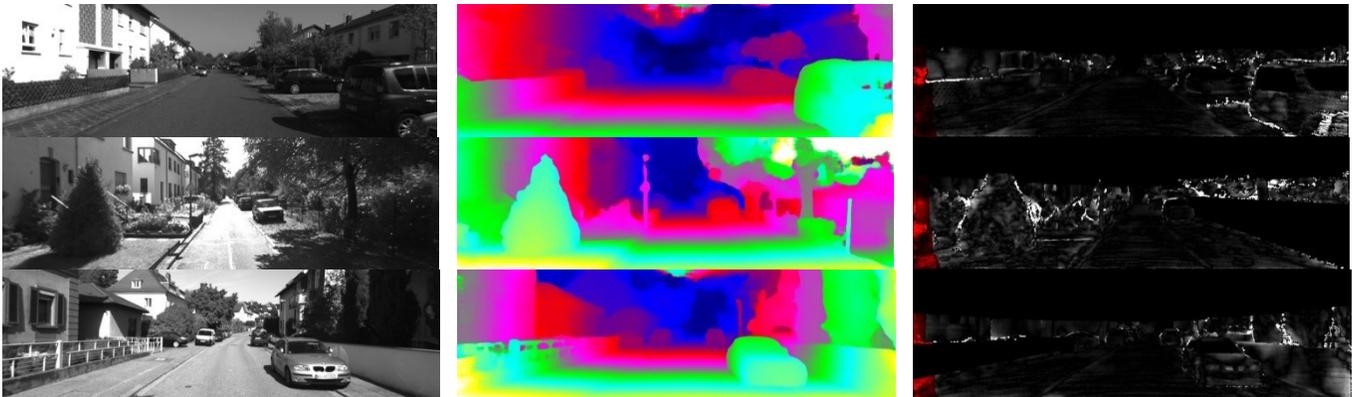

**Figure 6: KITTI 2012 test data qualitative results.** From left: left stereo input image, disparity prediction, error map.

As shown in Figures 6 and 7, our method could predict dense and clean disparity maps. MSDC-Net benefited from the Dense block has strong features extraction ability, compared to other methods. Thus, MSDC-Net could obtain more robust results, even in ill-posed regions. Our approach outperforms previous deep learning methods, which produce noisy and inaccuracy disparity maps. For this reason, these algorithms do not use multi-scale feature extraction and fusion architecture. Moreover, the layers of these algorithms are usually ~~more~~ shallow but have more parameters; it maybe limits the performance in the matching task.

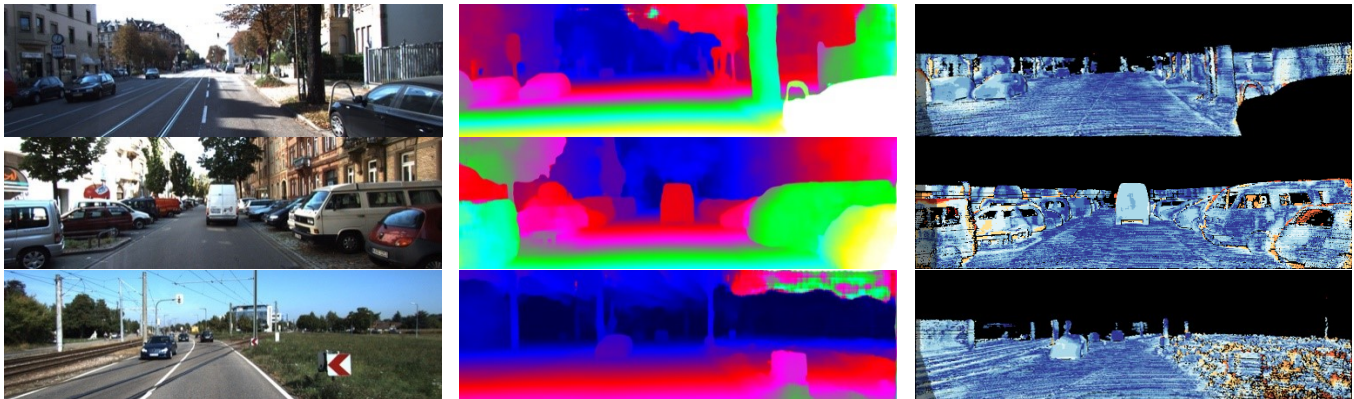

**Figure 7: KITTI 2015 test data qualitative results**. From left: left stereo input image, disparity prediction, error map.

**Table 2: Results on KITTI 2012 stereo benchmark.**

| Method | 2 pixels (%) | | 3 pixels (%) | | 4 pixels (%) | | 5 pixels (%) | | Avg-Noc | Avg-Noc |
|---|---|---|---|---|---|---|---|---|---|---|
| | Out-Noc | Out-All | Out-Noc | Out-All | Out-Noc | Out-All | Out-Noc | Out-All | | |
| L-ResMatch [15] | 3.64 | 5.06 | 2.27 | 3.40 | 1.76 | 2.67 | 1.50 | 2.26 | 0.7px | 1.0px |
| MC-CNN-acrt [16] | 3.90 | 5.45 | 2.37 | 3.63 | 1.90 | 2.85 | 1.64 | 2.39 | 0.7px | 0.9px |
| GC-NET [17] | 2.71 | 3.46 | 1.77 | 2.30 | 1.36 | 1.77 | 1.12 | 1.46 | 0.6px | 0.7px |
| PSMNet [18] | **2.44** | **3.01** | **1.49** | **1.89** | **1.12** | **1.42** | **0.90** | **1.15** | 0.5px | 0.6px |
| SsSMnet [19] | 3.34 | 4.24 | 2.30 | 3.00 | 1.82 | 2.39 | 1.53 | 2.01 | 0.7px | 0.8px |
| SGM-Net [20] | 3.60 | 5.15 | 2.29 | 3.50 | 1.82 | 2.39 | 1.60 | 2.36 | 0.7px | 0.9px |
| SPS-St [28] | 4.98 | 6.28 | 3.39 | 4.41 | 2.72 | 3.52 | 2.33 | 3.00 | 0.9px | 1.0px |
| Displets v2 [29] | 3.43 | 4.46 | 2.37 | 3.09 | 1.97 | 2.52 | 1.72 | 2.17 | 0.7px | 0.8px |
| PBCP [30] | 3.62 | 5.01 | 2.36 | 3.45 | 1.88 | 2.74 | 1.62 | 2.32 | 0.7px | 0.9px |
| MSDC-Net | 2.78 | 3.47 | 1.64 | 2.12 | 1.22 | 1.58 | 0.98 | 1.26 | **0.5px** | **0.6px** |

**Table 3: Results on KITTI 2015 stereo benchmark**.

| | All pixels (%) | | | 3 pixels (%) | | | |
|---|---|---|---|---|---|---|---|
| Method | D1-bg | D1-fg | D1-all | D1-bg | D1-fg | D1-all | Time (s) |
| L-ResMatch [15] | 2.72 | 6.95 | 3.42 | 2.35 | 5.76 | 2.91 | 48 |
| MC-CNN-acrt [16] | 2.89 | 8.88 | 3.89 | 2.48 | 7.64 | 3.33 | 67 |
| GC-NET [17] | 2.21 | 6.16 | 2.87 | 2.02 | 5.58 | 2.61 | 0.9 |
| PSMNet [18] | **1.86** | 4.62 | **2.32** | **1.71** | 4.31 | **2.14** | **0.41** |
| SsSMnet [19] | 2.70 | 6.92 | 3.40 | 2.46 | 6.13 | 3.06 | 0.8 |
| SGM-Net [20] | 2.66 | 8.64 | 3.66 | 2.23 | 7.44 | 3.09 | 67 |
| SPS-St [28] | 3.84 | 12.67 | 5.31 | 3.50 | 11.61 | 4.84 | 2 |
| Displets v2 [29] | 3.00 | 5.56 | 3.43 | 3.43 | 4.46 | 3.09 | 0.8 |
| PBCP [30] | 2.58 | 8.78 | 3.61 | 2.27 | 7.71 | 3.17 | 68 |
| MSDC-Net | 1.96 | **3.77** | 2.26 | 1.83 | **3.57** | 2.12 | 0.7 |

As shown in Table 2 and 3, Our model is better than GC-Net, Displets v2, SGM-Net and SsSMnet et al., which were reported by the KITTI evaluation server. Our method not only achieves state of the art results for both KITTI 2012 and 2015 benchmarks but also a little better than most competing approaches in the non-occluded region. Compared to other methods, our architecture more explicitly leverages different scale geometry by multi-scale fusion 2D convolution and multi-scale residual 3D convolution modules, resulting in an improvement in performance.

## 4. CONCLUSIONS

In this work, we propose a highly efficient network architecture for stereo matching. The proposed framework consists of two main modules: the multi-scale fusion 2D convolution module and the multi-scale residual 3D convolution module. The multi-scale fusion 2D convolution module incorporates different levels of feature maps to form a cost volume. The multi-scale residual 3D convolution module further learns to regularize the cost volume via repeated top-down/bottom-up processes. The proposed method has been verified through the different experiments. Experimental results show that the proposed method could predict a dense, clean and precise disparity map from image pairs. For future work, we are interested in exploring the potential of generative adversarial networks and explicit semantics to improve our disparity map prediction in the visual occlusion region.

## 5. ACKNOWLEDGMENTS

This work was supported in part by Natural Science Foundation of China (61420106007, 61671387 and 61871325).

MAV flight. In *Proceedings of IEEE International Conference on Robotics and Automation*.

# Columns on Last Page Should Be Made As Close As Possible to Equal Length

## Authors' background

| Your Name | Title* | Research Field | Personal website |
|---|---|---|---|
| Zhibo Rao | Ph.D. candidate | Deep Learning, Image processing | |
| Mingyi He | Professor | Neural network, Image processing | http://dianzi.nwpu.edu.cn/info/1317/8245.htm |
| Yuchao Dai | Professor | Deep Learning, Image processing | |
| Zhidong Zhu | master student | Deep Learning, Image processing | |
| Bo Li | Ph.D. candidate | Deep Learning, Image processing | |
| Renjie He | Research Fellow, Dr | Deep Learning, Image processing | |

*This form helps us to understand your paper better, **the form itself will not be published.**

**\*Title can be chosen from: master student, Phd candidate, assistant professor, lecture, senior lecture, associate professor, full professor**